\documentclass[conference]{IEEEtran}
\IEEEoverridecommandlockouts
\usepackage{cite}
\usepackage{amsmath,amssymb,amsfonts}
\usepackage{graphicx}
\usepackage{textcomp}
\usepackage{comment}
\usepackage{xcolor}
\usepackage[utf8]{inputenc}
\usepackage[T1]{fontenc}
\usepackage{hyperref}
\usepackage{url}
\usepackage{booktabs}
\usepackage{nicefrac}
\usepackage{microtype}
\usepackage{enumitem}
\usepackage{tcolorbox}
\usepackage{placeins}
\usepackage{float}
\usepackage{listings}
\usepackage{pdfpages}
\usepackage{longtable}
\usepackage{multirow}
\usepackage{verbatim}
\usepackage{stfloats}
\usepackage[export]{adjustbox}
\usepackage{placeins}
\usepackage{float}
\usepackage{tabularx}
\usepackage{pifont}
\usepackage{lipsum}
\usepackage{colortbl}
\usepackage{hyphenat}
\usepackage{array}
\usepackage{wrapfig}
\usepackage{fancyvrb}
\usepackage{rotating}

\usepackage{algorithm}
\usepackage[noend]{algpseudocode}

\definecolor{swecream}{HTML}{EFFEFF}
\definecolor{issueborder}{HTML}{15071A}
\definecolor{issuefill}{HTML}{F6F8FA}
\definecolor{envfill}{HTML}{F2F9FF}
\definecolor{envborder}{HTML}{123C7C}
\definecolor{agentfill}{HTML}{F9F3F3}
\definecolor{agentborder}{HTML}{9B0A0A}
\definecolor{goldpatchborder}{HTML}{FABB00}
\definecolor{goldpatchfill}{HTML}{FFF7E1}

\DefineVerbatimEnvironment{CodeVerbatim}{Verbatim}{
  formatcom={\color{black}},
  fontsize=\small,
  fontfamily=\ttdefault,
  fontseries=\mddefault,
  fontshape=\updefault,
  fillcolor=\color{white},
  framerule=0pt,
}

\usepackage{rotating} 

\usepackage{pifont}
\newcommand{\cmark}{\textcolor{green!60!black}{\ding{51}}}
\newcommand{\xmark}{\textcolor{red}{\ding{55}}}

\newtcolorbox{observationbox}[1][]{
        colback=envfill,
        colbacktitle=envfill,
        colframe=envborder,
        arc=5pt,
        fontupper=\small,
        fonttitle=\bfseries\color{black},
        boxrule=0.5mm,
        boxsep=1mm,
        width=\linewidth,
        breakable,
        title={Observation \hfill #1},
        rounded corners,
        toptitle=0.7mm,
        bottomtitle=0.7mm
}
\newtcolorbox{goldpatchbox}[1][]{
        colback=goldpatchfill,
        colbacktitle=goldpatchfill,
        colframe=goldpatchborder,
        arc=5pt,
        fontupper=\small,
        fonttitle=\bfseries\color{black},
        boxrule=0.5mm,
        boxsep=1mm,
        width=\linewidth,
        breakable,
        title={\twemoji{1f6a9} Flag Captured \hfill #1},
        rounded corners,
        toptitle=0.7mm,
        bottomtitle=0.7mm
}
\newtcolorbox{issuebox}[1][]{
        colback=issuefill,
        colbacktitle=issuefill,
        colframe=issueborder,
        arc=5pt,
        fontupper=\small,
        fonttitle=\bfseries\color{black},
        boxrule=0.5mm,
        boxsep=1mm,
        width=\linewidth,
        breakable,
        title={Issue \hfill #1},
        rounded corners,
        toptitle=1mm
}
\newtcolorbox{agentbox}[1][]{
        colback=agentfill,
        colbacktitle=agentfill,
        colframe=agentborder,
        arc=5pt,
        fontupper=\small,
        fonttitle=\bfseries\color{black},
        boxrule=0.5mm,
        boxsep=1mm,
        width=\linewidth,
        breakable,
        title={EnIGMA \hfill #1},
        rounded corners,
        toptitle=1mm,
        lower separated=false
}
\newtcolorbox{fileviewerbox}[1]{
        enhanced,
        breakable,
        boxrule = 1.5pt,
        fontupper = \small,
        fonttitle = \bf\color{black},
        arc = 5pt,
        rounded corners,
        colframe = black,
        colbacktitle = swecream,
        colback = swecream,
        title = #1,
        left=4pt 
}
\newtcolorbox{promptbox}[1]{
    enhanced,
    breakable,
    boxrule=1pt,  
    fontupper=\small,
    fonttitle=\bfseries\color{black},
    arc=3pt,  
    rounded corners,
    colframe=black,
    colbacktitle=swecream,
    colback=swecream,
    title=#1,
    left=2mm,  
    right=2mm,  
    top=1mm,  
    bottom=1mm  
}

\def\BibTeX{{\rm B\kern-.05em{\sc i\kern-.025em b}\kern-.08em
    T\kern-.1667em\lower.7ex\hbox{E}\kern-.125emX}}
\begin{document}

\title{Rethinking Evaluation of Multiple Sclerosis (MS) Lesion Segmentation Models}

\author{
\IEEEauthorblockN{Abdul Basit\textsuperscript{*}, Ashir Rashid\textsuperscript{*}, Muhammad Abdullah Hanif\textsuperscript{*}, Muhammad Shafique}
\IEEEauthorblockA{\textit{eBRAIN Lab, Division of Engineering}, \textit{New York University (NYU) Abu Dhabi}, Abu Dhabi, UAE\\
\{abdul.basit, ar7789, mh6117, muhammad.shafique\}@nyu.edu}
\thanks{\textsuperscript{*} Authors have equal contributions.}
\vspace{-10pt}
}


\maketitle

\begin{abstract}
Multiple Sclerosis (MS) is a chronic autoimmune disease that can significantly reduce the quality of life of a patient. Existing treatment options can only help slow down the progression of the disease. Therefore, early detection and precise monitoring of disease progression are important. Deep learning offers state-of-the-art models for detecting and segmenting MS lesions in brain MRI scans. However, most of these models are evaluated using the Dice score, without accounting for lesion-wise detection and segmentation performance or other metrics that quantify model performance in cases that are complex or confusing for human annotators, or in cases that are essential for disease detection and progression monitoring. In this paper, we highlight the need to rethink the evaluation of MS lesion segmentation models. In this context, we first present problem fingerprinting in detail to highlight what neurologists look for in brain MRI scans for MS detection and progression monitoring, and which metrics are required to properly quantify model performance in these contexts. Additionally, we present an analysis of state-of-the-art models on two open-source datasets using these metrics to highlight their usability for real-world deployment in hospitals.
\end{abstract}






\section{Introduction}

MS is an autoimmune condition in which the immune system attacks the protective covering, called the myelin sheath, that insulates nerve fibers (axons) in the brain and spinal cord, as well as the nerves that connect to the eyes~\cite{ghasemi2016multiple}. Myelin damage causes symptoms such as muscle weakness, tingling, numbness, or pain in the upper or lower limbs, vision problems, and memory issues~\cite{ghasemi2016multiple}. 
MS affects around 2.3~million people across the globe, and as of today, there is no cure for the disease, and the possible treatment options can only help in slowing down the progression and help the patient live a better life~\cite{haki2024review}. 
In this context, early detection and precise progression analysis of MS can play a vital role in facilitating timely neurologist intervention and providing appropriate treatment that can slow down the disease progression. 

MRI is the primary imaging test for detecting MS and monitoring its progression over time. This is essentially because MS-typical lesions appear as hyperintensities in T2-weighted and T2-FLAIR MRI scans~\cite{wattjes2015evidence}. Out of the two, T2-FLAIR (especially 3D FLAIR) is usually preferred as the core sequence for MS because in the T2-FLAIR sequence, lesions stay bright while Cerebrospinal Fluid is suppressed (appears dark), making it easier for clinicians and radiologists to detect, annotate, and longitudinally compare white-matter lesions~\cite{wattjes2015evidence}. Conventionally, the identification and annotation of lesions in MRI scans are performed manually by radiologists, which is a very tedious and time-consuming task and is thereby prone to errors and inter-observer variability. Various deep learning models have been proposed in the literature for automatic detection and segmentation of lesions in MRI scans~\cite{nnU-Net, SegResNet, SwinUNETR}. However, most works focus on the aggregate, image-level segmentation performance of the models (typically quantified using metrics such as Dice Similarity Coefficient (DSC), 95th percentile HD95, Average Symmetric Surface Distance (ASSD), and Positive Predictive Value (PPV)) along with lesion-wise detection performance, usually quantified using metrics such as lesion detection sensitivity (recall) and lesion detection precision (PPV), as summarized in Table~\ref{tab:ms_lesion_metrics_matrix}. \textit{Although these metrics provide a high-level view of a model's overall performance, they fail to highlight its usability across different scenarios, such as early MS detection and precise progression monitoring, as discussed in the motivational analysis below.}

\begin{table*}[t]
\caption{Presence (\cmark) or absence (\xmark) of common evaluation metric families in MS lesion segmentation papers.}
\label{tab:ms_lesion_metrics_matrix}
 \resizebox{1.0\textwidth}{!}{
\centering
\scriptsize
\setlength{\tabcolsep}{3pt}
\begin{tabular}{ll*{18}{p{1.1cm}}}
\toprule
& & \multicolumn{8}{c}{Voxel-wise / image-level metrics} 
& \multicolumn{6}{c}{Lesion-wise detection metrics} 
& \multicolumn{3}{c}{Lesion-wise seg. metrics} 
& \multicolumn{1}{c}{Size-aware} \\
\cmidrule(lr){3-10}
\cmidrule(lr){11-16}
\cmidrule(lr){17-19}
\cmidrule(lr){20-20}

RefID &
Year &
Dice / Jaccard / IoU &
Boundary dist.
(HD / HD95 / ASSD) &
Voxel sens.
(recall) &
Voxel prec.
(PPV) &
Voxel spec. &
Accuracy &
Volume error
(abs / rel) &
Volume corr.
(Pearson / ICC) &
Lesion sens.
(recall) &
Lesion prec.
(PPV) &
Lesion F1 &
FP lesions
per scan &
Lesion count
error / bias &
Lesion-level
ROC / FROC &
Lesion-wise
Dice &
Lesion-wise
distance &
Lesion-wise
vol. error &
Size- /
load-stratified
metrics \\
\midrule

\cite{Johnston1996SegmentationOMAV} &
1996 &
\cmark &  
\xmark &  
\xmark &  
\xmark &  
\xmark &  
\xmark &  
\xmark &  
\xmark &  
\xmark &  
\xmark &  
\xmark &  
\xmark &  
\xmark &  
\xmark &  
\xmark &  
\xmark &  
\xmark &  
\xmark \\  

\cite{Geremia2010SpatialDFAW} &
2010 &
\xmark &  
\xmark &  
\cmark &  
\cmark &  
\xmark &  
\xmark &  
\xmark &  
\xmark &  
\xmark &  
\xmark &  
\xmark &  
\xmark &  
\xmark &  
\xmark &  
\xmark &  
\xmark &  
\xmark &  
\xmark \\  

\cite{Abdullah2012MultiSectionalVTAX} &
2012 &
\cmark &  
\cmark &  
\cmark &  
\cmark &  
\xmark &  
\xmark &  
\cmark &  
\xmark &  
\cmark &  
\cmark &  
\xmark &  
\xmark &  
\xmark &  
\xmark &  
\xmark &  
\xmark &  
\xmark &  
\xmark \\  

\cite{Brosch2016Deep3CQ} &
2016 &
\cmark &  
\xmark &  
\xmark &  
\xmark &  
\xmark &  
\xmark &  
\cmark &  
\xmark &  
\cmark &  
\xmark &  
\xmark &  
\xmark &  
\xmark &  
\xmark &  
\xmark &  
\xmark &  
\xmark &  
\cmark \\  

\cite{Egger2016MRIFLAT} &
2016 &
\cmark &  
\xmark &  
\cmark &  
\xmark &  
\xmark &  
\xmark &  
\cmark &  
\cmark &  
\xmark &  
\xmark &  
\xmark &  
\xmark &  
\cmark &  
\xmark &  
\xmark &  
\xmark &  
\xmark &  
\cmark \\  

\cite{Jain2016TwoTPM} &
2016 &
\cmark &  
\xmark &  
\xmark &  
\xmark &  
\xmark &  
\xmark &  
\cmark &  
\cmark &  
\cmark &  
\cmark &  
\cmark &  
\xmark &  
\xmark &  
\xmark &  
\xmark &  
\xmark &  
\xmark &  
\cmark \\  

\cite{Valverde2017ImprovingAMAR} &
2017 &
\cmark &  
\xmark &  
\cmark &  
\cmark &  
\xmark &  
\xmark &  
\cmark &  
\cmark &  
\cmark &  
\xmark &  
\xmark &  
\xmark &  
\xmark &  
\cmark &  
\xmark &  
\xmark &  
\xmark &  
\xmark \\  

\cite{Carass2017LongitudinalMSAQ} &
2017 &
\cmark &  
\cmark &  
\cmark &  
\cmark &  
\xmark &  
\xmark &  
\cmark &  
\cmark &  
\cmark &  
\xmark &  
\xmark &  
\xmark &  
\cmark &  
\xmark &  
\xmark &  
\xmark &  
\xmark &  
\xmark \\  

\cite{Birenbaum2017MultiviewLCAB} &
2017 &
\cmark &  
\xmark &  
\xmark &  
\cmark &  
\xmark &  
\xmark &  
\xmark &  
\cmark &  
\cmark &  
\xmark &  
\xmark &  
\xmark &  
\xmark &  
\xmark &  
\xmark &  
\xmark &  
\xmark &  
\xmark \\  

\cite{Khastavaneh2017NeuralNLAY} &
2017 &
\cmark &  
\xmark &  
\cmark &  
\cmark &  
\cmark &  
\xmark &  
\xmark &  
\xmark &  
\xmark &  
\xmark &  
\xmark &  
\xmark &  
\xmark &  
\xmark &  
\xmark &  
\xmark &  
\xmark &  
\xmark \\  

\cite{Aslani2018MultibranchCNP} &
2018 &
\cmark &  
\cmark &  
\xmark &  
\cmark &  
\xmark &  
\xmark &  
\cmark &  
\cmark &  
\cmark &  
\xmark &  
\xmark &  
\xmark &  
\xmark &  
\xmark &  
\xmark &  
\xmark &  
\xmark &  
\xmark \\  

\cite{Commowick2018ObjectiveEOAP} &
2018 &
\cmark &  
\cmark &  
\cmark &  
\cmark &  
\cmark &  
\xmark &  
\xmark &  
\xmark &  
\cmark &  
\cmark &  
\cmark &  
\xmark &  
\xmark &  
\xmark &  
\xmark &  
\xmark &  
\xmark &  
\cmark \\  

\cite{Gabr2019BrainALC} &
2019 &
\cmark &  
\xmark &  
\xmark &  
\xmark &  
\xmark &  
\xmark &  
\xmark &  
\cmark &  
\cmark &  
\xmark &  
\xmark &  
\xmark &  
\cmark &  
\xmark &  
\xmark &  
\xmark &  
\xmark &  
\cmark \\  

\cite{Weeda2019ComparingLSZ} &
2019 &
\cmark &  
\xmark &  
\cmark &  
\xmark &  
\cmark &  
\xmark &  
\xmark &  
\cmark &  
\xmark &  
\xmark &  
\xmark &  
\xmark &  
\xmark &  
\xmark &  
\xmark &  
\xmark &  
\xmark &  
\xmark \\  

\cite{Zhang2019MultipleSLB} &
2019 &
\cmark &  
\xmark &  
\cmark &  
\cmark &  
\xmark &  
\xmark &  
\cmark &  
\xmark &  
\cmark &  
\xmark &  
\xmark &  
\xmark &  
\xmark &  
\xmark &  
\xmark &  
\xmark &  
\xmark &  
\xmark \\  

\cite{Zhang2019RSANetRSBP} &
2019 &
\cmark &  
\xmark &  
\xmark &  
\xmark &  
\xmark &  
\xmark &  
\xmark &  
\xmark &  
\xmark &  
\xmark &  
\xmark &  
\xmark &  
\xmark &  
\xmark &  
\xmark &  
\xmark &  
\xmark &  
\xmark \\  

\cite{McKinley2019SimultaneousLAAH} &
2019 &
\cmark &  
\xmark &  
\xmark &  
\xmark &  
\xmark &  
\xmark &  
\xmark &  
\cmark &  
\cmark &  
\cmark &  
\cmark &  
\xmark &  
\xmark &  
\xmark &  
\xmark &  
\xmark &  
\xmark &  
\xmark \\  

\cite{Cerri2020ACMBA} &
2020 &
\cmark &  
\xmark &  
\cmark &  
\cmark &  
\xmark &  
\xmark &  
\xmark &  
\cmark &  
\xmark &  
\xmark &  
\xmark &  
\xmark &  
\xmark &  
\xmark &  
\xmark &  
\xmark &  
\xmark &  
\cmark \\  

\cite{Denner2020SpatiotemporalLFBK} &
2020 &
\cmark &  
\xmark &  
\xmark &  
\xmark &  
\xmark &  
\xmark &  
\cmark &  
\xmark &  
\cmark &  
\cmark &  
\xmark &  
\xmark &  
\xmark &  
\xmark &  
\xmark &  
\xmark &  
\xmark &  
\xmark \\  

\cite{Kaur2020StateoftheArtSTBC} &
2020 &
\xmark &  
\xmark &  
\xmark &  
\xmark &  
\xmark &  
\xmark &  
\xmark &  
\xmark &  
\xmark &  
\xmark &  
\xmark &  
\xmark &  
\xmark &  
\xmark &  
\xmark &  
\xmark &  
\xmark &  
\xmark \\  

\cite{Zhang2021ALLNetAIO} &
2021 &
\cmark &  
\xmark &  
\cmark &  
\cmark &  
\xmark &  
\xmark &  
\xmark &  
\cmark &  
\cmark &  
\xmark &  
\cmark &  
\xmark &  
\xmark &  
\cmark &  
\xmark &  
\xmark &  
\xmark &  
\xmark \\  

\cite{Ma2021MultipleSLAI} &
2021 &
\xmark &  
\xmark &  
\xmark &  
\xmark &  
\xmark &  
\xmark &  
\xmark &  
\xmark &  
\xmark &  
\xmark &  
\xmark &  
\xmark &  
\xmark &  
\xmark &  
\xmark &  
\xmark &  
\xmark &  
\xmark \\  

\cite{Ansari2021MultipleSLAD} &
2021 &
\cmark &  
\xmark &  
\cmark &  
\cmark &  
\xmark &  
\xmark &  
\cmark &  
\cmark &  
\cmark &  
\xmark &  
\xmark &  
\xmark &  
\xmark &  
\xmark &  
\xmark &  
\xmark &  
\xmark &  
\xmark \\  

\cite{McKinley2021SimultaneousLABF} &
2021 &
\cmark &  
\xmark &  
\xmark &  
\xmark &  
\xmark &  
\xmark &  
\xmark &  
\xmark &  
\cmark &  
\cmark &  
\cmark &  
\xmark &  
\xmark &  
\xmark &  
\xmark &  
\xmark &  
\xmark &  
\xmark \\  

\cite{Cerri2022AnOTR} &
2022 &
\cmark &  
\xmark &  
\xmark &  
\xmark &  
\xmark &  
\xmark &  
\cmark &  
\cmark &  
\xmark &  
\xmark &  
\xmark &  
\xmark &  
\xmark &  
\xmark &  
\xmark &  
\xmark &  
\xmark &  
\xmark \\  

\cite{Krishnamoorthy2022FrameworkTSE} &
2022 &
\cmark &  
\xmark &  
\cmark &  
\cmark &  
\cmark &  
\cmark &  
\xmark &  
\xmark &  
\xmark &  
\xmark &  
\xmark &  
\xmark &  
\xmark &  
\xmark &  
\xmark &  
\xmark &  
\xmark &  
\xmark \\  

\cite{SadeghiBakhi2022MultipleSLD} &
2022 &
\cmark &  
\xmark &  
\xmark &  
\xmark &  
\xmark &  
\xmark &  
\cmark &  
\xmark &  
\cmark &  
\xmark &  
\xmark &  
\xmark &  
\xmark &  
\xmark &  
\xmark &  
\xmark &  
\xmark &  
\xmark \\  

\cite{D2022NoiseICAE} &
2022 &
\cmark &  
\xmark &  
\cmark &  
\cmark &  
\xmark &  
\xmark &  
\xmark &  
\xmark &  
\xmark &  
\xmark &  
\xmark &  
\xmark &  
\xmark &  
\xmark &  
\xmark &  
\xmark &  
\xmark &  
\xmark \\  

\cite{Hitziger2022TriplanarUWI} &
2022 &
\cmark &  
\xmark &  
\cmark &  
\cmark &  
\xmark &  
\xmark &  
\xmark &  
\xmark &  
\cmark &  
\cmark &  
\cmark &  
\xmark &  
\xmark &  
\xmark &  
\xmark &  
\xmark &  
\xmark &  
\xmark \\  

\cite{Hamad2022UsingCNAF} &
2022 &
\cmark &  
\xmark &  
\cmark &  
\cmark &  
\xmark &  
\xmark &  
\xmark &  
\xmark &  
\xmark &  
\xmark &  
\xmark &  
\xmark &  
\xmark &  
\xmark &  
\xmark &  
\xmark &  
\xmark &  
\xmark \\  

\cite{Rondinella2023BoostingMSA} &
2023 &
\cmark &  
\xmark &  
\cmark &  
\cmark &  
\cmark &  
\cmark &  
\xmark &  
\xmark &  
\xmark &  
\xmark &  
\xmark &  
\xmark &  
\xmark &  
\xmark &  
\xmark &  
\xmark &  
\xmark &  
\xmark \\  

\cite{Wu2023CoactSegLFF} &
2023 &
\cmark &  
\cmark &  
\xmark &  
\xmark &  
\xmark &  
\xmark &  
\xmark &  
\xmark &  
\cmark &  
\xmark &  
\cmark &  
\xmark &  
\xmark &  
\xmark &  
\xmark &  
\xmark &  
\xmark &  
\cmark \\  

\cite{Rondinella2023EnhancingMSAC} &
2023 &
\cmark &  
\xmark &  
\cmark &  
\cmark &  
\xmark &  
\xmark &  
\xmark &  
\xmark &  
\xmark &  
\xmark &  
\xmark &  
\xmark &  
\xmark &  
\xmark &  
\xmark &  
\xmark &  
\xmark &  
\xmark \\  

\cite{Liu2023MultipleSLAM} &
2023 &
\cmark &  
\xmark &  
\cmark &  
\xmark &  
\xmark &  
\xmark &  
\xmark &  
\xmark &  
\xmark &  
\xmark &  
\xmark &  
\xmark &  
\xmark &  
\xmark &  
\xmark &  
\xmark &  
\xmark &  
\xmark \\  

\cite{Hindsholm2023ScannerALAN} &
2023 &
\cmark &  
\xmark &  
\cmark &  
\cmark &  
\xmark &  
\xmark &  
\xmark &  
\xmark &  
\cmark &  
\cmark &  
\cmark &  
\xmark &  
\xmark &  
\xmark &  
\xmark &  
\xmark &  
\xmark &  
\cmark \\  

\cite{Amaludin2023TowardMAK} &
2023 &
\cmark &  
\xmark &  
\cmark &  
\cmark &  
\xmark &  
\xmark &  
\cmark &  
\cmark &  
\cmark &  
\xmark &  
\xmark &  
\xmark &  
\xmark &  
\xmark &  
\xmark &  
\xmark &  
\xmark &  
\xmark \\  

\cite{Zhang2023TowardsAAAK} &
2023 &
\cmark &  
\xmark &  
\cmark &  
\cmark &  
\xmark &  
\xmark &  
\xmark &  
\cmark &  
\cmark &  
\xmark &  
\xmark &  
\xmark &  
\xmark &  
\xmark &  
\xmark &  
\xmark &  
\xmark &  
\xmark \\  

\cite{Rosa2024ConsensusOAJ} &
2024 &
\cmark &  
\xmark &  
\cmark &  
\cmark &  
\xmark &  
\xmark &  
\cmark &  
\cmark &  
\cmark &  
\xmark &  
\xmark &  
\xmark &  
\cmark &  
\xmark &  
\xmark &  
\xmark &  
\xmark &  
\cmark \\  

\cite{Balan2024DiagnosisOML} &
2024 &
\xmark &  
\xmark &  
\cmark &  
\cmark &  
\xmark &  
\cmark &  
\xmark &  
\xmark &  
\xmark &  
\xmark &  
\xmark &  
\xmark &  
\xmark &  
\xmark &  
\xmark &  
\xmark &  
\xmark &  
\xmark \\  

\cite{Rondinella2024ICPR2CBD} &
2024 &
\cmark &  
\xmark &  
\xmark &  
\xmark &  
\xmark &  
\xmark &  
\xmark &  
\xmark &  
\xmark &  
\xmark &  
\xmark &  
\xmark &  
\xmark &  
\xmark &  
\xmark &  
\xmark &  
\xmark &  
\xmark \\  

\cite{Rokuss2024LongitudinalSOX} &
2024 &
\cmark &  
\cmark &  
\xmark &  
\xmark &  
\xmark &  
\xmark &  
\xmark &  
\xmark &  
\cmark &  
\cmark &  
\cmark &  
\xmark &  
\xmark &  
\xmark &  
\xmark &  
\xmark &  
\xmark &  
\xmark \\  

\cite{Wiltgen2024LSTAIADN} &
2024 &
\cmark &  
\cmark &  
\cmark &  
\cmark &  
\xmark &  
\xmark &  
\xmark &  
\xmark &  
\cmark &  
\cmark &  
\cmark &  
\xmark &  
\xmark &  
\xmark &  
\xmark &  
\xmark &  
\xmark &  
\cmark \\  

\cite{Basaran2024SegHeDSOV} &
2024 &
\cmark &  
\xmark &  
\xmark &  
\xmark &  
\xmark &  
\xmark &  
\xmark &  
\cmark &  
\cmark &  
\cmark &  
\cmark &  
\xmark &  
\xmark &  
\xmark &  
\xmark &  
\xmark &  
\xmark &  
\cmark \\  

\cite{Basaran2024SegHeDSOAJ} &
2024 &
\cmark &  
\xmark &  
\xmark &  
\xmark &  
\xmark &  
\xmark &  
\xmark &  
\cmark &  
\cmark &  
\cmark &  
\cmark &  
\xmark &  
\xmark &  
\xmark &  
\xmark &  
\xmark &  
\xmark &  
\cmark \\  

\cite{Dereskewicz2025ANCH} &
2025 &
\cmark &  
\xmark &  
\xmark &  
\cmark &  
\xmark &  
\xmark &  
\cmark &  
\xmark &  
\cmark &  
\xmark &  
\cmark &  
\xmark &  
\xmark &  
\xmark &  
\xmark &  
\xmark &  
\xmark &  
\cmark \\  

\cite{Cetin2025EnhancingPIW} &
2025 &
\cmark &  
\xmark &  
\xmark &  
\xmark &  
\xmark &  
\xmark &  
\xmark &  
\xmark &  
\xmark &  
\xmark &  
\xmark &  
\xmark &  
\xmark &  
\xmark &  
\xmark &  
\xmark &  
\xmark &  
\xmark \\  

\cite{Dereskewicz2025FLAMeSARG} &
2025 &
\cmark &  
\xmark &  
\cmark &  
\cmark &  
\xmark &  
\xmark &  
\cmark &  
\xmark &  
\cmark &  
\xmark &  
\cmark &  
\xmark &  
\xmark &  
\xmark &  
\xmark &  
\xmark &  
\xmark &  
\cmark \\  

\cite{Guarnera2025MSLesSegBAAU} &
2025 &
\cmark &  
\cmark &  
\cmark &  
\cmark &  
\xmark &  
\xmark &  
\cmark &  
\xmark &  
\cmark &  
\xmark &  
\xmark &  
\xmark &  
\xmark &  
\xmark &  
\xmark &  
\xmark &  
\xmark &  
\xmark \\  

\cite{Darestani2025} &
2025 &
\cmark &  
\cmark &  
\cmark &  
\cmark &  
\xmark &  
\cmark &  
\xmark &  
\xmark &  
\xmark &  
\xmark &  
\xmark &  
\xmark &  
\xmark &  
\xmark &  
\xmark &  
\xmark &  
\xmark &  
\xmark \\  

\bottomrule
\end{tabular}
}
\end{table*}

\subsection{Motivation behind Rethinking Evaluation of MS Lesion Segmentation Models}

\begin{figure*}[t]
\centering
\includegraphics[width=1\linewidth]{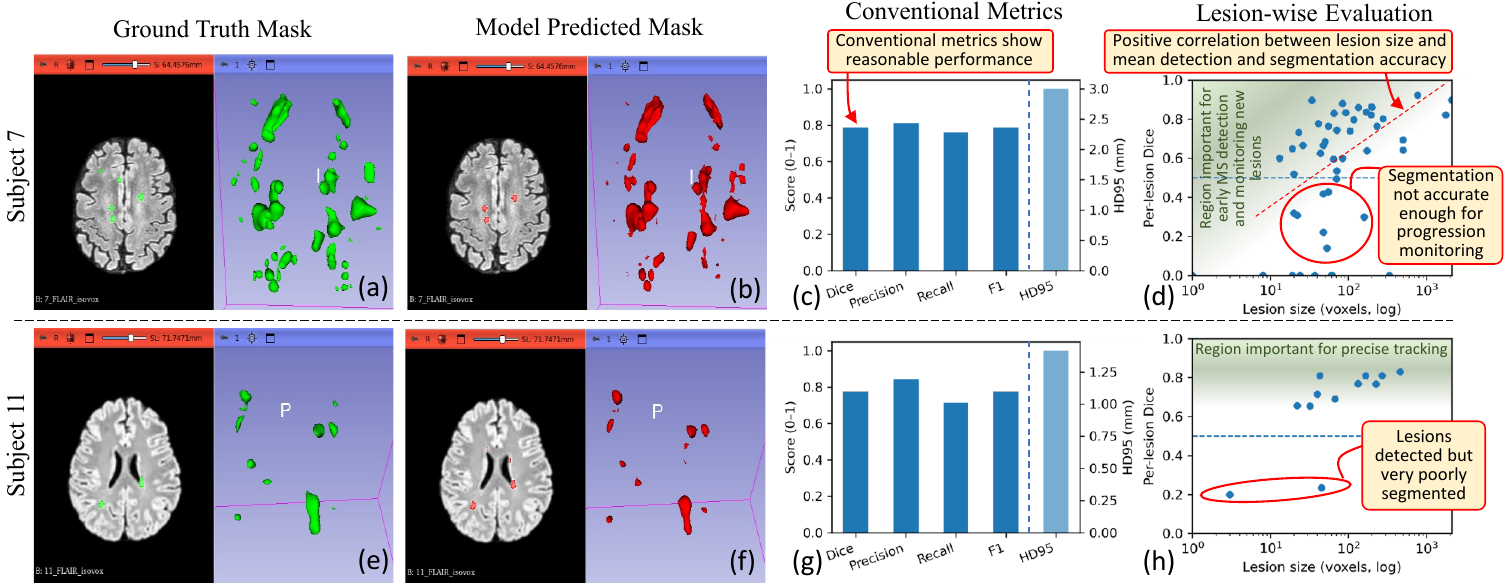}
\caption{Performance of the nnU-Net model, trained on MSSEG-1 dataset, on two validation samples (Subject~7 and Subject~11) from MSSEG-1. (a) and (e) show the ground-truth masks, while (b) and (f) illustrate the model-predicted masks. The model’s performance on the two samples, in terms of conventional metrics, is shown in (c) and (g), respectively. The respective lesion-wise Dice results are presented in (d) and (h).}
\label{fig:motivation}
\end{figure*}

In MS, clinical decision-making depends on lesion count, lesion appearance/disappearance, and lesion-wise volume changes rather than voxel-wise overlap alone. 
Voxel-based metrics inherently bias performance evaluation toward larger lesions in the dataset, thereby masking systematic failures in detecting and segmenting small but clinically relevant lesions, and also downplaying false-positive detections due to this same size-related bias. 
Fig.~\ref{fig:motivation} summarizes the performance of the nnU-Net model, trained on MSSEG-1 dataset, on two validation samples from MSSEG-1. As can be observed from the figure, the conventional metrics highlight reasonable model performance, i.e., around 0.78 DSC score. However, lesion-wise evaluation, which explicitly detects, matches, and analyzes the accuracy of individual lesions, enables disentangling detection errors from overall instance-level segmentation quality errors. 
From the results in the figure, we can draw the following conclusions:

\begin{enumerate}[leftmargin=12pt]
    \item Performing lesion-wise evaluation and mapping lesions in Dice vs. lesion size space reveals performance characteristics that are not captured by conventional aggregate metrics. 
    \item Lesion detection failures are disproportionately concentrated in small lesions, which are underrepresented in voxel-wise metrics. 
    \item Lesion-wise Dice scores correlate weakly with voxel-wise, instance-level Dice for small lesions but more strongly for large lesions. In general, the lesion-wise Dice scores are lower for small lesions and tend to approach the aggregate value for large lesions. 
    \item Almost all lesions are segmented imperfectly, with at least some volumetric prediction error. 
\end{enumerate}

In summary, lesion-wise evaluation provides complementary and necessary information beyond conventional segmentation scores. 
Therefore, a standardized lesion-wise evaluation framework is necessary to provide a more complete, clinically meaningful, and interpretable assessment of MS lesion segmentation models. 

\subsection{Our Novel Contributions}

\begin{enumerate}[leftmargin=12pt]
    \item In this paper, we highlight that, as aggregate evaluation metrics fail to quantify a model's performance across lesions properly and to provide assurance of acceptable segmentation quality across all lesions, there is a need to rethink how MS lesion segmentation models are evaluated. Additionally, if a model cannot detect and segment lesions accurately, its capabilities should be quantified in a way that allows radiologists to understand the types and extent of errors the automatic segmentation model can make, thereby enabling them to efficiently scan MRI-annotation pairs for such errors. 
    \item We highlight that lesion-wise evaluation of segmentation models is essential for understanding the variability in model performance across lesions, as only such evaluation can reveal whether a model has a high probability of segmenting a lesion with very low accuracy or is more likely to miss a specific type of lesion than others.
    \item We propose an evaluation framework based on fingerprinting the MS lesion segmentation problem from the perspectives of neurologists and radiologists. The evaluation framework mainly focuses on lesion-wise assessment of models to identify variations in performance across lesions as well as to identify limitations in detecting and segmenting small, medium, and large lesions in a brain MRI scan. 
    \item We present a detailed evaluation of five state-of-the-art MS lesion segmentation models on two open-source datasets, and highlight their limitations for real-world deployment. 
\end{enumerate}

\section{Lesion-wise Evaluation Framework}
{
\subsection{Problem Fingerprinting}
\label{sec:fingerprinting}
An MS lesion segmentation model is typically trained to annotate all types of lesions in a brain MRI scan, regardless of their size and shape characteristics or the patient's disease stage. The disease stage can usually define the presence or absence of lesions with specific characteristics or sizes~\cite{la2018shallow}. Therefore, based on disease stage, the scans can broadly be classified into the following categories, as illustrated in Fig.~\ref{fig:stages}. 


\begin{enumerate}[leftmargin=12pt]
    \item MRI with no lesion, corresponding to no disease
    \item MRI with small lesions, corresponding to a very early disease stage
    \item MRI with small and medium-sized lesions, corresponding to intermediate (moderate) disease stage
    \item MRI with small, medium, and large lesions, corresponding to the advanced stage.
\end{enumerate}


Usually, in the early stages, detection of the disease is critical, and then progression monitoring. Even a small number of false positives can lead to incorrect MS detection, and false negatives can lead to misdiagnosis. Therefore, precise lesion detection is essential for accurate disease diagnosis, specifically for detection at an early stage~\cite{li2018multi}. 
Additionally, precise segmentation of existing lesions, regardless of size, is critical for tracking disease progression over time. 
Studies have shown that identification and precise measurement specifically of small lesions is critical for evaluating treatment effects~\cite{wang1997survey}. 
Over-segmentation of a lesion can lead to a false perception of disease progression and under-segmentation can lead to delayed treatment. 

\begin{figure*}[t]
\centering
\includegraphics[width=0.90\linewidth]{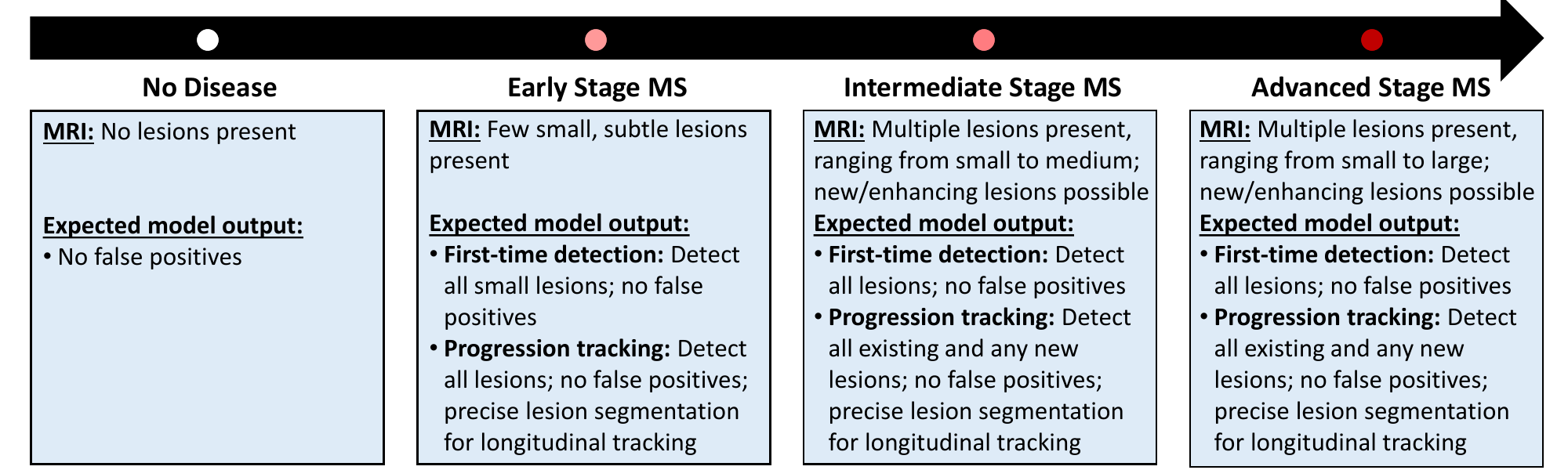}
\caption{MS disease stages and the corresponding expectations for an automated lesion segmentation model.}
\label{fig:stages}
\end{figure*}

\subsection{Metrics Identification}

The problem fingerprinting phase highlights the importance of correctly detecting and segmenting individual lesions for accurate, early disease detection and precise monitoring over time. In this context, both lesion-wise detection and segmentation metrics are important for evaluating a model's overall performance. Therefore, we propose to use the lesion-level Dice score to compute the overlap (similarity) between a lesion in the ground truth and its corresponding detected lesion in the predicted mask. The lesion-wise Dice provides a perspective of average performance per lesion and also highlights worst-case scenarios. 
Additionally, from the detection perspective, we propose to use lesion-wise detection metrics such as precision, recall, and F1.
Also, we categorize lesions having different sizes based on thresholds defined in voxels. These lesion stratification allows us to evaluate the model performance on lesion-level rather than globally comparing the models.

    
    

\color{black}
\subsection{Evaluation Pipeline}
\label{sec:eval_pipeline}

\begin{figure*}[t]
\centering
\includegraphics[width=0.90\linewidth]{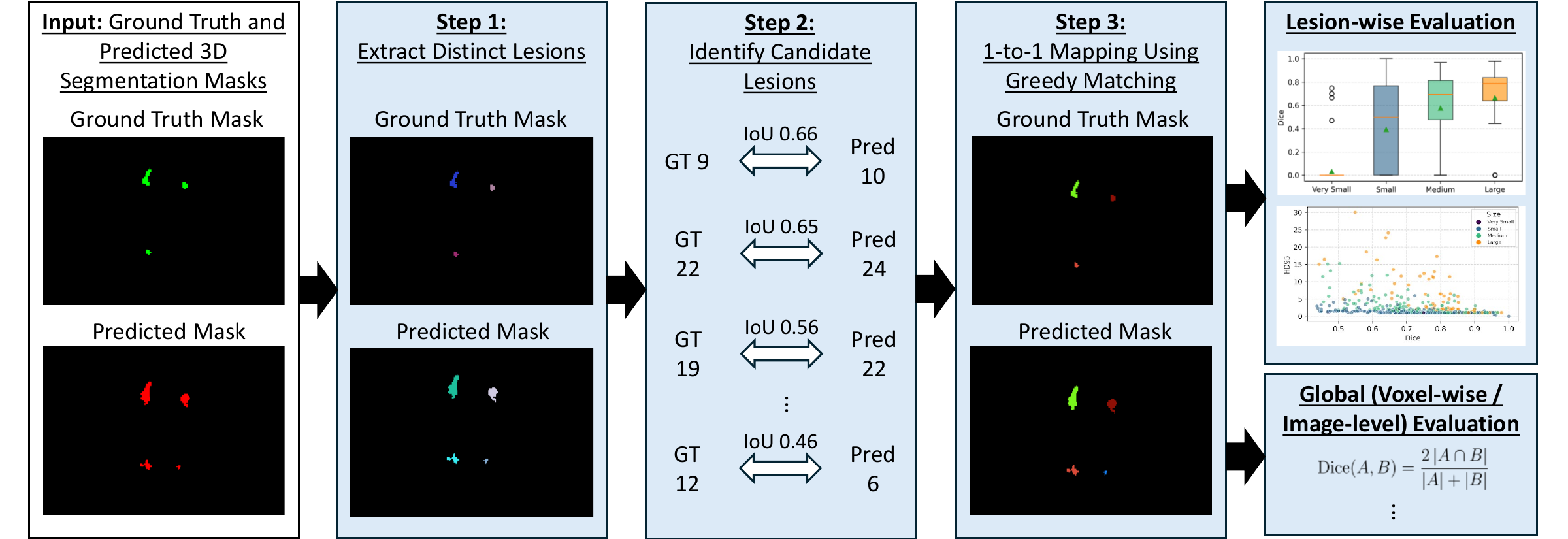}
\caption{Overview of our proposed evaluation pipeline.}
\label{fig:Frame}
\end{figure*}

Figure~\ref{fig:Frame} presents an overview of our proposed three-step evaluation pipeline. 
The pipeline evaluates segmentation performance at the lesion level. It first transforms voxel masks into distinct lesion objects via Connected Component Analysis. Then, it establishes a strict one-to-one correspondence between Ground Truth and Predicted lesions using a greedy Intersection-over-Union (IoU) matching strategy. This method ensures robust calculation of detection (TP, FP, FN) and segmentation metrics for matched pairs, providing a comprehensive assessment of model performance. The pseudo-code of the pipeline is presented in Algo.~\ref{algo:metrics}, and the details of each step are described below. 

\subsubsection{Detect and Extract Lesions (Fig.~\ref{fig:Frame}, Step~1)}
In this step, binary segmentation masks are decomposed into individual lesions using 3D connected component analysis using scipy's ndimage module. This transformation from voxel-wise maps to discrete objects enables instance-level evaluation.

\subsubsection{Identify Candidate Pairs (Fig.~\ref{fig:Frame}, Step~2)}
To identify plausible matches, we iterate through all GT-Predicted pairs using a two-step filter. First, disjoint pairs are efficiently discarded via Bounding Box intersection checks. Second, for overlapping pairs, we compute the Intersection over Union (IoU). Pairs exceeding a threshold $\tau$ are stored as candidates $(G, P, \text{IoU})$, allowing for distinct lesions to overlap with multiple targets initially.

\subsubsection{Greedy Matching Strategy (Fig.~\ref{fig:Frame}, Step~3)}
This phase resolves ambiguities to enforce a strict one-to-one mapping. Candidate pairs are sorted by IoU in descending order and known as potential matches. A pair $(G, P)$ is confirmed as a Match only if both lesions remain unmatched. Once confirmed, they are locked and removed from the pool. This greedy approach prioritizes the highest quality overlaps first, preventing double-counting and ensuring metric integrity.

\subsubsection{Size Stratified Evaluation}
Post-evaluation, performance is analyzed across different lesion sizes to ensure the model detects both substantial and subtle abnormalities. Lesions are categorized based on their voxel count into predefined bins: Very Small (0–10 voxels), Small (10–100 voxels), Medium (100–400 voxels), and Large (>400 voxels). These bins correspond to increasing lesion volumes in mm$^3$, improving clinical interpretability. Metrics such as Dice and HD95 are then aggregated for each size category, highlighting potential biases or failures in detecting smaller lesions.

    
    

\begin{algorithm}
\caption{Lesion- and Instance-Level Metric Computation}
\label{algo:metrics}
\small
\begin{algorithmic}[1]
\State \textbf{Input:} Ground-truth 3D mask $M_{GT}$, predicted 3D mask $M_{Pred}$, IoU threshold $\tau$
\State \textbf{Output:} Lesion-wise metrics $R_{Lesion}$, instance-level metrics $R_{Instance}$, size-stratified results $D_{Stratified}$

\State \textit{// Extract connected lesions}
\State $L_{GT} \gets \text{FindConnectedComponents}(M_{GT})$
\State $L_{Pred} \gets \text{FindConnectedComponents}(M_{Pred})$

\State \textit{// Generate candidate lesion pairs}
\State $C \gets [\ ]$
\For{$G \in L_{GT}$}
    \For{$P \in L_{Pred}$}
        \If{$\neg \text{BBoxOverlap}(G, P)$}
            \State \textbf{continue}
        \EndIf
        \State $iou \gets \text{ComputeIoU}(G, P)$
        \If{$iou > \tau$}
            \State $C.\text{append}((G,P,iou))$
        \EndIf
    \EndFor
\EndFor

\State \textit{// Greedy one-to-one matching}
\State Sort $C$ by decreasing $iou$
\State $S_{GT} \gets \emptyset$
\State $S_{Pred} \gets \emptyset$
\State $M \gets \emptyset$
\State $R_{Lesion} \gets \emptyset$

\For{$(G,P,iou) \in C$}
    \If{$G.id \notin S_{GT}$ \textbf{and} $P.id \notin S_{Pred}$}
        \State $S_{GT}.\text{add}(G.id)$
        \State $S_{Pred}.\text{add}(P.id)$
        \State $M.\text{add}((G,P))$
        \State $R_{Lesion}.\text{append}(\text{ComputeLesionMetrics}(G,P))$
    \EndIf
\EndFor

\State \textit{// Compute instance-level metrics}
\State $R_{Instance} \gets \text{ComputeInstanceMetrics}(L_{GT}, L_{Pred}, M)$

\State \textit{// Stratify matched lesions by ground-truth size}
\State $D_{Stratified} \gets \emptyset$
\For{$r \in R_{Lesion}$}
    \State $bin \gets \text{Categorize}(r.V_{GT})$
    \State $D_{Stratified}[bin].\text{append}(r)$
\EndFor

\State \Return $R_{Lesion}, R_{Instance}, D_{Stratified}$
\end{algorithmic}
\end{algorithm}

\section{Experimental Setup}
\label{sec:experimental_setup}

\subsection{Datasets}

We evaluate our framework on two publicly available MS lesion segmentation datasets: MSSEG-1 and MSLesSeg.

\textbf{MSSEG-1} is a multi-centre challenge dataset of 3D brain MRI scans with manually annotated white-matter lesions on T2-FLAIR images \cite{Carass2017LongitudinalMSAQ,Commowick2018ObjectiveEOAP}. Each subject is provided with voxel-wise lesion masks generated by multiple raters; following common practice, we use the consensus masks released by the challenge organisers as ground truth. We use the official training split for model development and reserve the validation subjects for evaluation and figures
throughout this paper. The evaluation is conducted on 24 held-out test cases.

\textbf{MSLesSeg} is a recently released large-scale dataset of brain MRI scans with lesion annotations on T2-FLAIR images \cite{Guarnera2025MSLesSegBAAU}. Compared to MSSEG-1, MSLesSeg contains a larger variety of scanners and acquisition protocols, and covers a broader range of lesion loads. The evaluation is conducted on 22 held-out test cases.

For both datasets, we resample all volumes to a common voxel spacing in millimetres and rigidly align them to a standard orientation. Intensities are z-score normalised per scan using non-zero brain voxels. All experiments are conducted on the T2-FLAIR modality only, which is the clinical workhorse sequence for MS lesion assessment. 

We set the threshold parameter $\tau = 0.35$, which is determined via hyperparameter tuning on a synthetic 2D dataset designed to mimic lesion characteristics. This value is kept fixed across all experiments.
\subsection{Models and Implementation}

We consider three fully supervised 3D convolutional segmentation networks and three publicly available tools that are widely used in clinical research:

\begin{itemize}
    \item \textbf{nnU-Net}~\cite{nnU-Net}: a self-configuring U-Net based architecture that has become a strong baseline across medical segmentation tasks.
    \item \textbf{SegResNet}~\cite{SegResNet}: a residual encoder--decoder network with anisotropic down-sampling, implemented
          in MONAI.
    \item \textbf{SwinUNETR}\cite{SwinUNETR}: a hybrid transformer--U-Net architecture with a Swin Transformer encoder.
    \item \textbf{SAMSEG}~\cite{Cerri2020ACMBA}: a classical Bayesian segmentation tool for brain tissues and lesions.
    \item \textbf{LST-AI}~\cite{Wiltgen2023LSTAIADY}: a deep learning extension of the Lesion Segmentation Tool (LST) for automated MS lesion segmentation.

\end{itemize}

All deep neural network models (nnU-Net, SegResNet, and SwinUNETR) are implemented and trained using the MONAI framework on top of PyTorch. For each dataset, we train a separate model instance per architecture. Where applicable, we initialise hyperparameters from publicly available reference implementations and adapt only image size, batch size, and number of training epochs to fit GPU memory constraints.

\subsection{Training Protocol}
We trained \textbf{nnU-Net}, \textbf{Swin-UNETR}, and \textbf{SegResNet} using the nnU-Net V2 framework on two datasets. The first dataset, \textbf{MSLesSeg}, comprises 93 training samples with a single FLAIR input channel, while the second, \textbf{MSSEG-1}, includes 23 training samples with five input channels (FLAIR, T1, T2, PD, T1CE) used in a 5-fold cross-validation setup. All images were preprocessed by resampling to an isotropic $1.0 \times 1.0 \times 1.0$ mm spacing and applying channel-wise Z-score normalization based on foreground intensity statistics.

Each model was trained for 1000 epochs using the \texttt{3d\_fullres} configuration with a batch size of 2 and a patch size of $128^3$ voxels. We utilized the standard nnU-Net training pipeline, which employs a combination of Dice and Cross-Entropy loss and a poly learning rate decay schedule. To mitigate overfitting, we applied extensive online data augmentation, including random rotations, scaling, gaussian noise, gaussian blur, brightness and contrast adjustments, low-resolution simulation, gamma correction, and mirroring.

\subsection{Evaluation Metrics}

For evaluation, we used the proposed evaluation pipeline presented in Section~\ref{sec:eval_pipeline}. For lesion-wise metrics, we considered Dice, 95th percentile HD95, F1-score, Precision, and Recall. For voxel-wise evaluation, we focused on the Dice similarity coefficient and HD95 to assess the volumetric overlap and boundary alignment accuracy of each model, respectively.

\begin{table*}[ht]
\centering
\caption{Size-Stratified Performance of State-of-the-Art models on MSSEG-1 and MSLesSeg Datasets}
\label{tab:size_stratified_global_metrics_2dp_f1}

\setlength{\tabcolsep}{2.0pt}      
\renewcommand{\arraystretch}{1.05} 

\resizebox{\textwidth}{!}{%
\begin{tabular}{@{}ll*{20}{c}@{}}
\toprule
Dataset & Model
& \multicolumn{5}{c}{Very Small}
& \multicolumn{5}{c}{Small}
& \multicolumn{5}{c}{Medium}
& \multicolumn{5}{c}{Large} \\
\cmidrule(lr){3-7}\cmidrule(lr){8-12}\cmidrule(lr){13-17}\cmidrule(lr){18-22}
& & Dice & HD95 & Prec. & Recall & F1
  & Dice & HD95 & Prec. & Recall & F1
  & Dice & HD95 & Prec. & Recall & F1
  & Dice & HD95 & Prec. & Recall & F1 \\
\midrule

\multirow{5}{*}{\textbf{MSSEG-1}} & LSTAI
& 0.06 & 1.00 & 0.04 & 0.09 & 0.05
& 0.31 & 1.32 & 0.57 & 0.45 & 0.50
& 0.29 & 2.00 & 0.66 & 0.38 & 0.48
& 0.41 & 8.67 & 0.75 & 0.49 & 0.59 \\
& SAMSEG
& 0.02 & 1.09 & 0.01 & 0.03 & 0.01
& 0.08 & 1.82 & 0.37 & 0.12 & 0.18
& 0.29 & 2.73 & 0.69 & 0.42 & 0.52
& 0.54 & 10.18 & 0.76 & 0.69 & 0.73 \\
& SegResNet
& 0.05 & 0.94 & 0.07 & 0.07 & 0.07
& 0.43 & 1.23 & 0.73 & 0.58 & 0.65
& 0.61 & 1.80 & 0.88 & 0.77 & 0.82
& 0.62 & 7.06 & 0.74 & 0.80 & 0.77 \\
& SwinUNETR
& 0.08 & 0.97 & 0.07 & 0.11 & 0.08
& 0.41 & 1.26 & 0.66 & 0.56 & 0.61
& 0.52 & 1.64 & 0.79 & 0.66 & 0.72
& 0.58 & 7.00 & 0.76 & 0.76 & 0.76 \\
& nnU-Net
& 0.07 & 1.00 & 0.08 & 0.10 & 0.09
& 0.44 & 1.30 & 0.73 & 0.61 & 0.66
& 0.58 & 2.02 & 0.82 & 0.75 & 0.78
& 0.68 & 6.83 & 0.87 & 0.82 & 0.85 \\

\midrule
\multirow{5}{*}{\textbf{MSLesSeg}} & LSTAI
& 0.00 & $-$    & 0.00 & 0.00 & $-$
& 0.27 & 1.59 & 0.45 & 0.41 & 0.43
& 0.42 & 2.22 & 0.72 & 0.59 & 0.65
& 0.60 & 5.34 & 0.87 & 0.77 & 0.82 \\
& SAMSEG
& 0.01 & 1.41 & 0.00 & 0.01 & 0.01
& 0.06 & 2.00 & 0.40 & 0.08 & 0.14
& 0.24 & 2.89 & 0.63 & 0.34 & 0.44
& 0.51 & 7.14 & 0.81 & 0.69 & 0.74 \\
& SegResNet
& 0.03 & 1.03 & 0.05 & 0.05 & 0.05
& 0.36 & 1.29 & 0.63 & 0.48 & 0.54
& 0.56 & 2.09 & 0.82 & 0.72 & 0.77
& 0.64 & 4.26 & 0.85 & 0.80 & 0.83 \\
& SwinUNETR
& 0.02 & 1.00 & 0.02 & 0.02 & 0.02
& 0.36 & 1.27 & 0.60 & 0.47 & 0.53
& 0.54 & 2.23 & 0.75 & 0.70 & 0.73
& 0.62 & 4.46 & 0.85 & 0.79 & 0.82 \\
& nnU-Net
& 0.03 & 1.04 & 0.06 & 0.05 & 0.05
& 0.39 & 1.29 & 0.71 & 0.54 & 0.61
& 0.58 & 2.62 & 0.86 & 0.77 & 0.82
& 0.72 & 4.29 & 0.92 & 0.88 & 0.90 \\

\bottomrule
\end{tabular}%
}
\end{table*}

\section{Results}

\subsection{General Model Performance Trends}

Table \ref{tab:size_stratified_global_metrics_2dp_f1} highlights a clear performance gap across lesion sizes, with all models achieving their highest Dice and F1 scores on "Large" lesions. Across both MSSEG-1 and MSLesSeg datasets, nnU-Net and SegResNet consistently outperform LSTAI and SAMSEG in the Medium and Large categories, particularly in terms of Precision and Recall. Interestingly, while LSTAI underperforms in overlap metrics for small lesions, it often maintains competitive HD95 values, reinforcing the trend that its detections are spatially well-localized. Notably, the MSLesSeg dataset appears more challenging for the "Very Small" category, with several models reporting near-zero Recall and F1 scores.

\subsection{Comparative Evaluation: nnU-Net on MSSEG-1 vs. MSLesSeg} When comparing nnU-Net on MSLesSeg (Figure \ref{fig:results_nnU-Net_mslesseg}) to the MSSEG-1 baseline (Figure \ref{fig:results_nnU-Net_msseg}), several shifts in metric distributions are observable. In Plot (B), median Dice scores for medium and large lesions are higher on MSLesSeg, while Plot (C) shows a lower, more concentrated HD95 distribution across most size categories. Plot (D) for MSSEG-1 reveals a higher density of outliers with low Dice scores and high HD95 values, whereas MSLesSeg results show a more linear inverse correlation between these metrics with fewer extreme spatial deviations.

The distribution of detection errors also varies between evaluations. Plot (E) indicates that for MSSEG-1, false positives are primarily "Very Small," while on MSLesSeg, the total false positive count is lower and the distribution mode shifts toward "Small" lesions. These trends are reflected in Plot (F), where MSSEG-1 shows higher inter-patient volatility in error proportions, whereas MSLesSeg displays more consistent true-positive counts relative to lesion volume.


\begin{figure*}[t]
\centering
\includegraphics[width=1\linewidth]{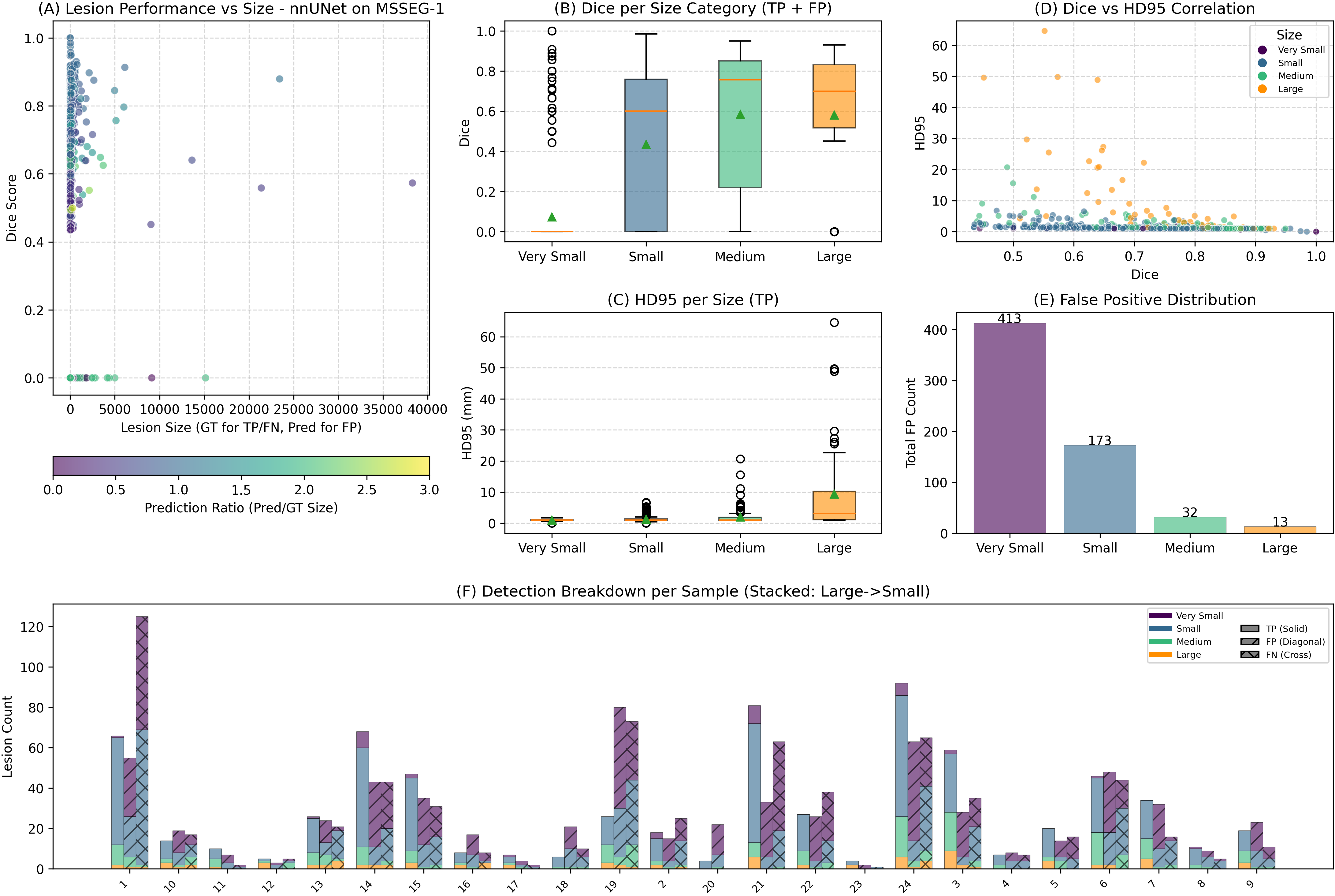}
\vspace{-10pt}
\caption{\textbf{nnU-Net Performance Evaluation on MSSEG-1} (A) \textit{Lesion Performance Scatter}: Correlates Dice scores with lesion size, where color indicates the Predicted Lesion Size to GT Lesion Size Ratio. (B) \textit{Volumetric Stratification (Dice)}: Distribution of Dice scores across size bins (Very Small to Large). (C) \textit{Volumetric Stratification (HD95)}: Distribution of HD95 across size bins. (D) \textit{Metric Correlation}: Relationship between Dice and HD95, colored by size category. (E) \textit{False Positive Analysis}: Count of false positive detections per size bin. (F) \textit{Detection Breakdown per Sample}: Stacked bar chart showing TP, FP, and FN counts for each validation case, stratified and colored by lesion size.\vspace{-10pt}}
\label{fig:results_nnU-Net_msseg}
\end{figure*}

\begin{figure*}[t]
\centering
\includegraphics[width=1\linewidth]{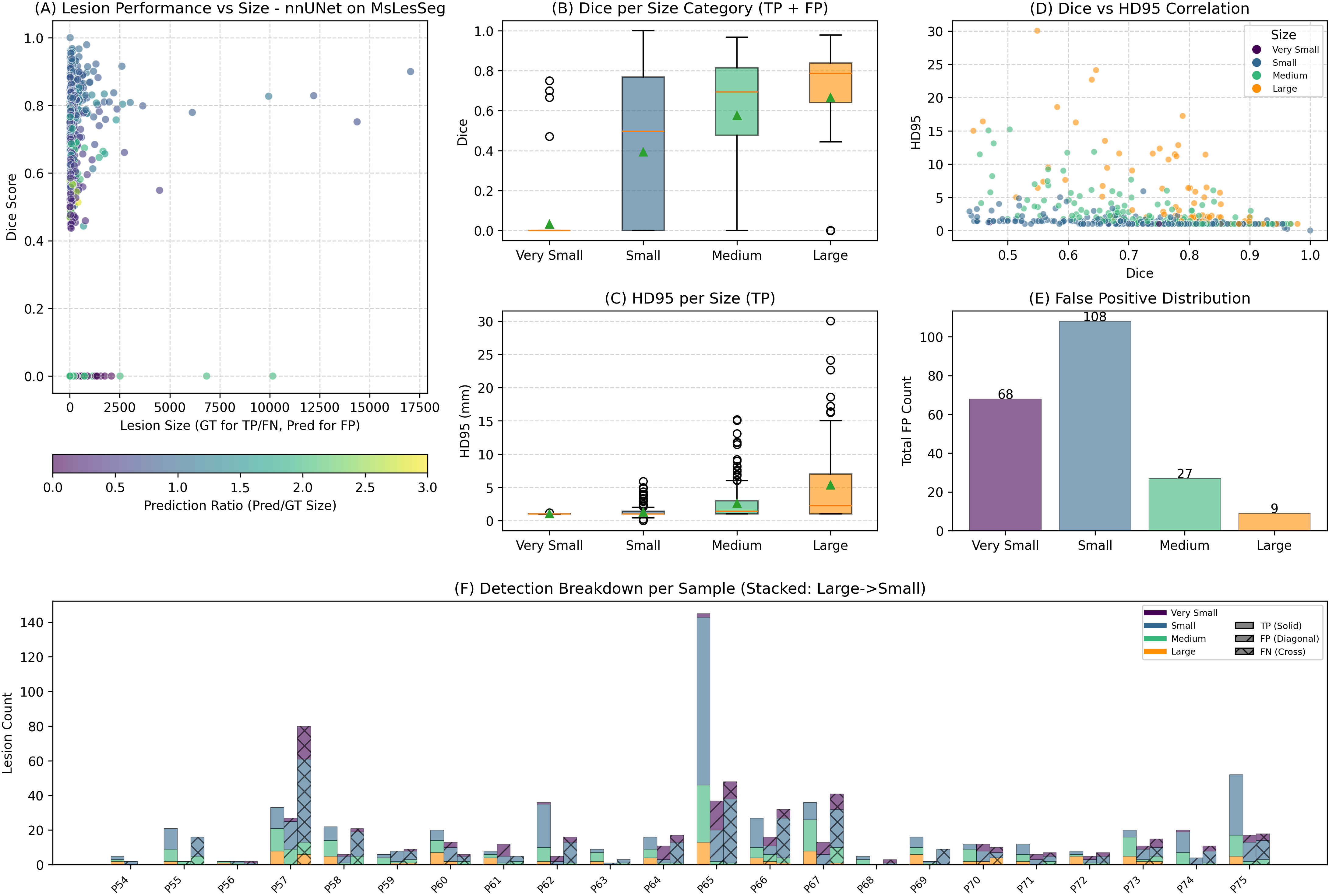}
\vspace{-10pt}
\caption{\textbf{nnU-Net Performance Evaluation on MSLesSeg}: The same structure as Figure \ref{fig:results_nnU-Net_msseg}\vspace{-10pt}}
\label{fig:results_nnU-Net_mslesseg}
\end{figure*}


\section{Conclusion}
In this paper, we highlight the need for evaluating the MS lesion segmentation models from the usability perspective that takes into account the criticality of different lesions and better highlight the types and extent of errors a model can make. We proposed to use lesion-wise metrics for model evaluation. Our analysis highlights that all state-of-the-art models lack in detecting and precisely segmenting small lesions. Our analysis highlights that all state-of-the-art models, including nnU-Net and SegResNet, exhibit a significant performance gap across lesion sizes, with a disproportionate concentration of detection failures and segmentation inaccuracies in small and very small lesions. While aggregate metrics often suggest high performance, our framework reveals that these models frequently miss clinically vital markers necessary for early MS detection and precise progression monitoring.

\vspace{-5pt}
}


\section*{Acknowledgment}
This research was partially funded by National Multiple Sclerosis Society (NMSS) under the ``LAMINATE: Longitudinal AI-based MS Lesion and Atrophy Segmentation Tool'' project and the NYUAD Center for Artificial Intelligence and Robotics (CAIR), funded by Tamkeen under the NYUAD Research Institute Award CG010.

\bibliographystyle{IEEEtran}
\bibliography{cite}

\end{document}